\documentclass[letterpaper]{article} 
\usepackage{aaai23}  
\usepackage{times}  
\usepackage{helvet}  
\usepackage{courier}  
\usepackage[hyphens]{url}  
\usepackage{graphicx} 
\usepackage{natbib}  
\usepackage{caption} 
\usepackage{algorithm}
\usepackage{algorithmic}
\usepackage{ulem}
\usepackage{booktabs}
\usepackage{multirow}
\usepackage{enumitem}
\usepackage{amssymb}
\usepackage{bbding}
\usepackage{svg}

\usepackage{newfloat}
\usepackage{listings}
\DeclareCaptionStyle{ruled}{labelfont=normalfont,labelsep=colon,strut=off} 
\floatstyle{ruled}
\newfloat{listing}{tb}{lst}{}
\floatname{listing}{Listing}
\title{ConTextual Masked Auto-Encoder for Dense Passage Retrieval}
\author{
    Xing Wu\textsuperscript{\rm 1,2,3 \equalcontrib}, Guangyuan Ma\textsuperscript{\rm 1,2 \equalcontrib}, Meng Lin\textsuperscript{\rm 1,2}, Zijia Lin\textsuperscript{\rm 3}, Zhongyuan Wang\textsuperscript{\rm 3}, Songlin Hu\textsuperscript{\rm 1,2}\thanks{Corresponding author.}
}
\affiliations{
    \textsuperscript{\rm 1} Institute of Information Engineering, Chinese Academy of Sciences \\
    \textsuperscript{\rm 2} School of Cyber Security, University of Chinese Academy of Sciences \\
    \textsuperscript{\rm 3} Kuaishou Technology

    \{wuxing, maguangyuan, linmeng, husonglin\}@iie.ac.cn, linzijia07@tsinghua.org.cn, wangzhongyuan@kuaishou.com
    
}

\usepackage{bibentry}

\begin{document}

\maketitle

\begin{abstract}
Dense passage retrieval aims to retrieve the relevant passages of a query from a large corpus based on dense representations (i.e., vectors) of the query and the passages.
Recent studies have explored improving pre-trained language models to boost dense retrieval performance.
This paper proposes CoT-MAE (ConTextual Masked Auto-Encoder), a simple yet effective generative pre-training method for dense passage retrieval.
CoT-MAE employs an asymmetric encoder-decoder architecture that learns to compress the sentence semantics into a dense vector through self-supervised and context-supervised masked auto-encoding.
Precisely, self-supervised masked auto-encoding learns to model the semantics of the tokens inside a text span, and context-supervised masked auto-encoding learns to model the semantical correlation between the text spans.
We conduct experiments on large-scale passage retrieval benchmarks and show considerable improvements over strong baselines, demonstrating the high efficiency of CoT-MAE. Our code is available at \url{https://github.com/caskcsg/ir/tree/main/cotmae}.
\end{abstract}

\section{Introduction}
Passage retrieval aims to retrieve the relevant passages of a query from a large corpus, which benefits many downstream applications, such as web search \cite{fan2021pre, guo2022semantic, lin2021pretrained}, question answering \cite{karpukhin2020dense, lee2020learning, zhu2021adaptive} and dialogue systems \cite{gao2022neural, yu2021few}.

\begin{figure*}
\centering
\includegraphics[width=18cm]{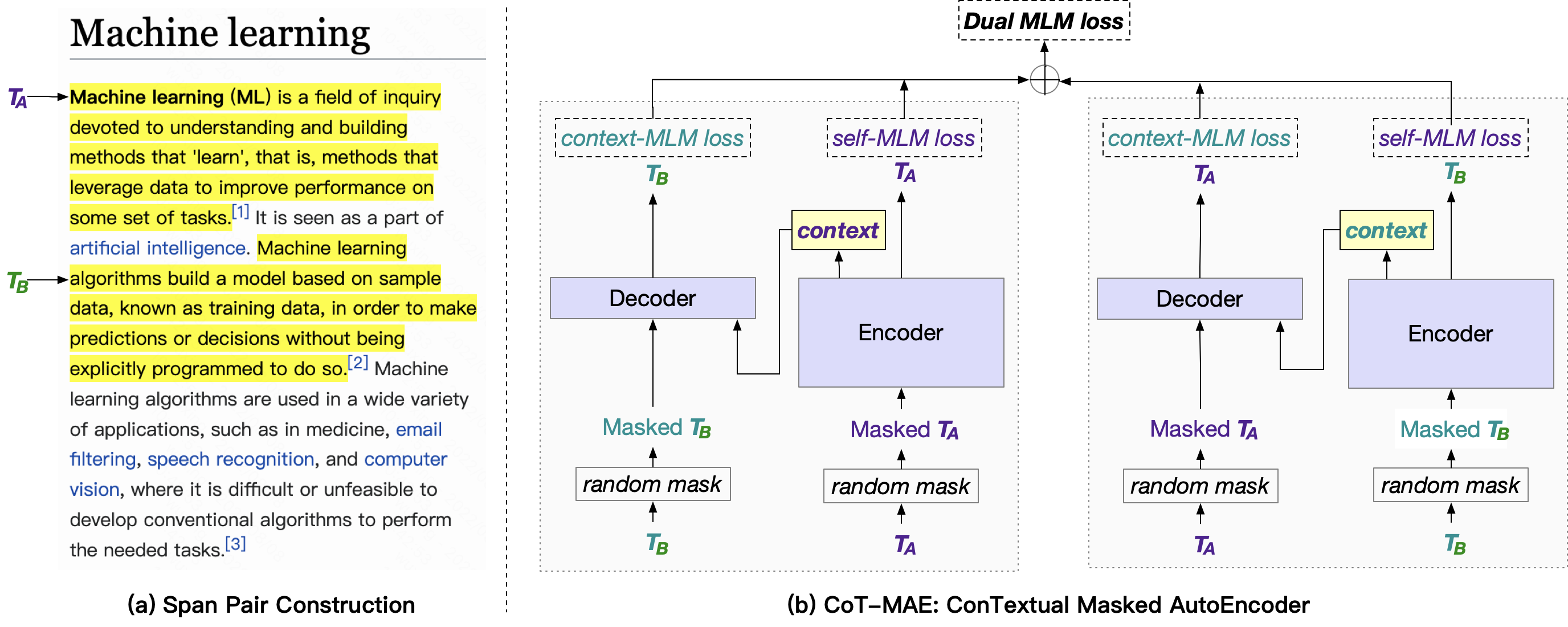}
\caption{CoT-MAE. (a) The process of span pair construction. We select two neighboring text spans $\mathbf{T_A}$ and $\mathbf{T_B}$ from a document with a sampling strategy to form a span pair. The two spans in a pair are each other's context. (b) The model design for CoT-MAE. We use an asymmetric encoder-decoder structure, with a deep encoder having enough parameters to learn good text representations modeling ability and a shallow decoder to assist the encoder in achieving this goal.}
\label{fig_cmae}
\end{figure*}
For a long time, sparse retrieval represented by BM25 \cite{robertson2009probabilistic} was the dominant retrieval method.
Recently, dense retrieval has received increasing attention with the development of pre-trained language models (PLM) \cite{devlin2018bert, liu2019roberta}.
Dense retrieval models are generally based on pre-trained language models with a siamese or dual-encoder architecture to encode queries and documents into low-dimensional vector space for effective search  \cite{hofstatter2021efficiently, humeau2019poly, xiong2020approximate, zhan2021optimizing, zhan2020repbert}. The relevances between queries and documents are calculated with cosine similarity or dot-product function in the vector space. 
Therefore, high-quality text representation based on PLM is crucial for dense passage retrieval.

DPR\cite{karpukhin2020dense} successfully shows that dense retrieval models can outperform BM25 methods.
Since then, some works have emerged to boost dense retrieval performance by improving the pre-training process tailored for dense retrieval. 
\cite{lu2021less, gao2021condenser, liu2022retromae} encourage the encoder to improve the text representation modeling ability through auxiliary self-supervised reconstruction tasks.
Auxiliary tasks usually utilize a weak decoder to reconstruct the masked text with the assistance of the text's vector from the encoder, which forces the encoder to provide better text representations. 
Although these works have been shown to be very effective and achieved some improvements in dense retrieval, \textit{they mainly focus on single-text internal modeling without considering contextual information.}
\cite{chang2020pre, gao2021unsupervised, ma2022pre} proposes multi-source and multi-granularity contrastive span prediction tasks to model the semantic similarity between relevant text spans during pre-training.
These discriminative tasks introduced in the pre-training improve dense retrieval and the trained retrievers achieve state-of-the-art performance.
Meanwhile, in multimodal representation learning \cite{li2021align, li2022blip}, generative pre-training tasks, such as image-based text decoding, have been proven effective, \textit{leaving us with a question of how well span-level context-grounded generative auxiliary task will boost the dense retrieval.}

To improve text representation modeling for dense passage retrieval, this paper proposes a novel pre-training method by further introducing semantic correlations between text spans in a generative manner.
When modeling, we jointly consider the semantics of the tokens inside a text span and the semantical correlation between the text spans.
As shown in Figure \ref{fig_cmae}-(a), we select two neighboring text spans $\mathbf{T_A}$ and $\mathbf{T_B}$ from a document with a sampling strategy to form a span pair.
Then, as shown in Figure \ref{fig_cmae}-(b), we use the masked auto-encoding to model the whole process.
First, we consider self-supervised masked auto-encoding, which reconstructs a masked text span only considering the unmasked tokens in the span on the encoder side.
As shown in the purple flow in Figure \ref{fig_cmae}-(b), we apply masking operations to $\mathbf{T_A}$ and $\mathbf{T_B}$ respectively, feeding the masked text to the encoder to reconstruct the original text with (\textit{self-})masked language modeling (MLM) objective.
Next, we consider context-supervised auto-encoding, which reconstructs a masked text span jointly considering the unmasked tokens in the span and its neighboring span embedding, i.e., its context, on the decoder side.
As shown in the green flow in Figure \ref{fig_cmae}-(b), on the decoder side, let's take $\mathbf{T_B}$ as an example.
We apply a mask operation to $\mathbf{T_B}$ and provide the masked $\mathbf{T_B}$ with its context, i.e., the encoded embedding of $\mathbf{T_A}$, to the decoder, jointly reconstructing the $\mathbf{T_B}$ with \textit{context-}MLM objective.
We call the whole process ConTextual Masked AutoEncoding (CoT-MAE) and uniformly optimize with the MLM loss.
Note that we only use the decoder in pre-training. During the dense retrieval fine-tuning, the decoder is no longer needed.

There are several benefits to our design.
First, CoT-MAE uses self-supervised and context-supervised masked auto-encoding tasks. The two tasks are linked by context representation and jointly optimized to learn better text representation modeling capabilities.
Second, the asymmetric encoder-decoder structure is very flexible. 
We use a deep encoder with adequate parameters to learn a good text representation modeling ability.
We use a shallow decoder without enough parameters to recover the masked tokens well, which results in a strong reliance on the context from the encoder and forces the encoder to learn to provide better text representation.
Finally, we can use asymmetric masking operations, that is, using different masking rates on the encoder and decoder sides.
Inspired by \cite{he2022masked, wettig2022should} that a large mask rate may benefit pre-training, we adopt up to 30\% and 45\% mask rate in CoT-MAE's encoder and decoder, respectively. A larger mask rate on the decoder side further increases the difficulty of context-supervised masked auto-encoding and forces the encoder side to acquire more powerful text encoding capabilities.

To verify the effectiveness of our proposed CoT-MAE, we conduct experiments on large-scale web search and open-domain QA benchmarks: MS-MARCO Passage Ranking \cite{nguyen2016ms}, TREC Deep Learning (DL) Track 2020 \cite{craswell2020overview}, and Natural Questions (NQ) \cite{kwiatkowski2019natural}. Experimental results show that CoT-MAE achieves considerable gains over competing baseline retrievers. In addition, we perform extensive ablation experiments to illustrate the soundness of the CoT-MAE design. 

Our contributions can be summarized as follows:
\begin{enumerate}
 \item[-] We propose a novel generative pre-training method CoT-MAE tailored for dense retrieval.
 \item[-] We design a data construction method and an asymmetric encoder-decoder structure for efficient pre-training.
 \item[-] Experiments show that CoT-MAE achieves considerable gains over competing retrievers on benchmarks.
\end{enumerate}

\section{Related Works}
DPR \cite{karpukhin2020dense} outperforms BM25 methods with dense retrieval models.
Since then, many works have emerged to boost dense retrieval performance, including techniques to improve pre-training and fine-tuning.
\paragraph{Pre-training tasks tailored for dense retrieval}
One category \cite{lu2021less, gao2021condenser, liu2022retromae} forces the encoder to provide better text representations with auxiliary self-supervised auto-encoding tasks. 
For example, \cite{lu2021less, gao2021condenser} proposes to perform the auto-encoding using a weak decoder, with restricted capacity and attention flexibility to push the encoder to provide better text representation.
\cite{liu2022retromae} proposes to apply asymmetric masking ratios to the encoder and the weak decoder. A sentence embedding from the encoder is combined with its aggressively masked version to reconstruct the original sentence by the decoder.
Similar to these works, our method adopts an asymmetric encoder and weak decoder architecture. 
Differently, we propose context-supervised auto-encoding, in which a masked version of a given text and its neighboring text's embeddings, i.e., its context, from the encoder are jointly fed into the decoder to reconstruct the given text.

The other category's works \cite{chang2020pre, gao2021unsupervised, ma2022pre} propose multi-source and multi-granularity contrastive span prediction tasks to resemble passage retrieval in pre-training.
\cite{chang2020pre} proposes three tasks: inverse cloze task (ICT), body first selection (BFS), and wiki link prediction (WLP). The three tasks exploit the document structure of Wikipedia pages to automatically generate contrastive pairs and pull closer relevant pairs while pushing away irrelevant ones.
Similarly, \cite{gao2021unsupervised} introduces a corpus-level contrastive span prediction loss to the pre-training process, with the hypothesis that spans from the same document are closer than those from different documents.
\cite{ma2022pre} generalizes the contrastive span prediction task to several levels of granularity, i.e., word-level, phrase-level, sentence-level and paragraph-level.
Differently, we introduce a generative non-contrastive contextual masked auto-encoding task via the decoder-side MLM reconstruction assisted by contextual embedding, which is more challenging but proved to be more effective.
\paragraph{Fine-tuning}
Many attempts have explored to improve fine-tuning peformance, such as mining hard negatives \cite{xiong2020approximate, zhan2021optimizing}, late interaction \cite{khattab2020colbert}, distill knowledge from a strong teacher \cite{lin2021batch, santhanam2021colbertv2}, query clustering \cite{hofstatter2021efficiently}, data augmentation \cite{qu2020rocketqa} and jointly optimize retriever and re-ranker \cite{ren2021rocketqav2, zhang2022hlatr, zhang2021adversarial} . 

For example, \cite{xiong2020approximate} proposes to construct hard negatives by searching the corpus with a periodic updated Approximate Nearest Neighbor (ANN) index, which has been proved very effective and adopted by the following methods. 
\cite{zhan2021optimizing} further utilizes fine-tuned dense retriever to improve the quality of mined hard negatives.
\cite{khattab2020colbert} proposed a late interaction that applies a MaxSim operation on the last hidden states of the encoder to model the fine-grained similarity between queries and documents. 
\cite{lin2021batch} distills from ColBERT’s MaxSim operator into a retriever, meanwhile \cite{santhanam2021colbertv2} proposes to distill from a stronger re-ranker to the ColBERT.
\cite{hofstatter2021efficiently} introduce an efficient topic-aware query and balanced margin sampling technique to improve the fine-tuning efficiency.
 \cite{qu2020rocketqa} combines three effective strategies to achieve good performance, i.e., cross-batch negatives, denoised hard negatives, and data augmentation. 
\cite{ren2021rocketqav2} introduce the dynamic listwise distillation by designing a unified listwise training approach to improve both the retriever and the re-ranker adaptively.
\cite{zhang2022hlatr} designs Hybrid List Aware Transformer Reranking (HLATR) as a subsequent reranking module to incorporate retrieval and reranking stage features.
\cite{zhang2021adversarial} present adversarial retriever-ranker, which consists of a dual-encoder retriever and a cross-encoder ranker to be jointly optimized according to a minimax adversarial objective.
As we focus on the improvement brought by pre-training, following \cite{gao2021condenser, gao2021unsupervised, ma2022pre}, we reuse the open source fine-tuning pipeline Tevatron \cite{gao2022tevatron} to evaluate the effectiveness of our pre-training method.


\section{Approach}
In this section, we first introduce the masked auto-encoder structure as preliminary knowledge. Then we introduce the data construction and model structure of CoT-MAE.

\subsection{Preliminary: Masked Auto-Encoding}
Textual Masked Auto-Encoding, i.e., BERT's MLM task, is trained on unlabeled data without requiring additional manual labeling.
Formally, given a text $\mathbf{T}$ with $n$ consequent tokens, we prepend a special token [CLS] to the beginning of the text as
\begin{equation}
    \mathbf{T} = \{t_0, t_1, ..., t_n\}
\end{equation}
, in which the $t_0$ denotes the [CLS] token.
We randomly select a certain percentage, i.e., mask rate, of tokens and replace them with special token [MASK].
Inspired by \cite{he2022masked, wettig2022should} that a large mask rate may be beneficial for pre-training, we employ a mask rate greater than the BERT's 15\% setting.
We denote the tokens replaced by [MASK] as $m(\mathbf{T} )$ and the rest tokens as $\mathbf{T} {\backslash m(\mathbf{T} )}$. 
$\mathbf{T} {\backslash m(\mathbf{T} )}$ is then passed through the encoder to recover $m(\mathbf{T} )$ with the masked language model(MLM) loss. 
We formulate this process as:
\begin{equation}
\mathcal{L}_{mlm}=-\sum_{t \in m(\mathbf{T})} \log p(t \mid \mathbf{T} {\backslash m(\mathbf{T} )})
\end{equation}
For the $l$-th transformer layer in the encoder or decoder, its outputs are the hidden states of the layer
\begin{equation}
    \mathbf{h}^{l} = \{h_0^{l}, h_1^{l}, ..., h_n^{l}\}
\end{equation}
. Usually, the hidden states of the [CLS] position in the last layer of the encoder,  i.e., $h_0^{last}$, will be used as the embedding representation of $\mathbf{T}$.

\subsection{CoT-MAE: ConTextual Masked Auto-Encoder}
CoT-MAE jointly learns to model the semantics of the tokens inside a text span and the semantical correlation between the text spans.

We first describe how to build training data from unlabeled documents, as shown in Figure \ref{fig_cmae}-(a).
Given a document, we use tools like NLTK to split it into text spans that do not exceed a maximum length.
Then we select two neighboring text spans $\mathbf{T_A}$ and $\mathbf{T_B}$ from a document with a sampling strategy to form a span pair. The two spans in a pair are each other's neighbor or context.
\paragraph{Sampling Strategies}
We use three sampling strategies, termed Near, Olap and Rand. 
The Near strategy samples two adjacent spans without overlapping to form a pair. The Olap strategy samples two adjacent spans with partially overlapping segments to form a pair.
The Rand strategy randomly samples two non-overlapping spans to form a pair.

Then, we introduce the model design for CoT-MAE, as shown in Figure \ref{fig_cmae}-(b).
We use an asymmetric encoder-decoder structure, with a strong deep encoder and a weak shallow decoder.
The deep encoder has enough parameters to learn good text representation modeling ability, and the shallow decoder is set to assist the encoder in achieving this goal.
As the CoT-MAE adopts a dual modeling process for a span pair, we will only introduce the left half of Figure \ref{fig_cmae}-(b) in detail. That is,  $\mathbf{T_A}$ is on the encoder side, and $\mathbf{T_B}$ is on the decoder side.
We adopt asymmetric masking operations, using different mask rates on the encoder and decoder sides.
We apply random mask operation to $\mathbf{T_A}$ on the encoder side. We denote the tokens replaced by [MASK] as $m_{enc}(\mathbf{T_A} )$ and the rest tokens as $\mathbf{T_A} {\backslash m_{enc}(\mathbf{T_A} )}$. 
Similarly, we apply another random mask operation to $\mathbf{T_B}$ on the decoder side. We denote the tokens replaced by [MASK] as $m_{dec}(\mathbf{ T_B} )$ and the rest tokens as $\mathbf{T_B} {\backslash m_{dec}(\mathbf{T_B} )}$. 

\paragraph{Self-supervised Pre-training}
On the encoder side, we reconstruct a text span only considering the unmasked tokens in the span.
The unmasked tokens $\mathbf{T_A} {\backslash m_{enc}(\mathbf{T_A} )}$ is passed through the encoder to recover $m_{enc}(\mathbf{T_A} )$ with the self-supervised masked language model(self-MLM) loss:
\begin{equation}
\mathcal{L}^{A}_{smlm}=-\sum_{t \in m(\mathbf{T_A})} \log p(t \mid \mathbf{T_A} {\backslash m(\mathbf {T_A} )})
\end{equation}
. The subscript ``smlm'' denotes the process is self-supervised, superscript $A$ denotes the self-supervised pre-training is applied on $T_A$.
\paragraph{Context-supervised Pre-training}
On the decoder side,  we reconstruct the other text span in the pair considering its unmasked tokens and its neighboring span embedding, i.e., its context embedding.
Specifically, for $\mathbf{T_B}$, its context embedding is the [CLS] hidden state of $\mathbf{T_A}$ from the encoder's last layer, also denoted as $h_0^{last}$.
We jointly feed  $\mathbf{T_B} {\backslash m_{dec}(\mathbf{T_B} )}$ and $h_0^{last}$ into the decoder to recover $m(\mathbf{T_B} )$ using the context-supervised masked language model loss:
\begin{equation}
\mathcal{L}^{AB}_{cmlm}=-\sum_{t \in m(\mathbf{T_B})} \log p(t \mid [h_0^{last}, \mathbf{T_B} {\backslash m(\mathbf {T_B} )]})
\end{equation}
. The subscript ``cmlm'' denotes the process is context-supervised, the superscript ``AB'' denotes that $\mathbf {T_B}$ uses $\mathbf {T_A}$ as context, and ``[]'' denotes concatenation operation.
Then, we add the losses from both the encoder and the decoder to get a summed loss:
\begin{equation}
\mathcal{L}^{AB} = \mathcal{L}^{A}_{smlm} + \mathcal{L}^{AB}_{cmlm}
\end{equation}
At the same time, there is also a dual case, $\mathbf {T_B}$ is on the encoder side and $\mathbf {T_A}$ is on the encoder side. The summed loss is:
\begin{equation}
\mathcal{L}^{BA} = \mathcal{L}^{B}_{smlm} + \mathcal{L}^{BA}_{cmlm}
\end{equation}
Finally, the total loss of our proposed CoT-MAE is:
\begin{equation}
\mathcal{L} = \mathcal{L}^{AB} + \mathcal{L}^{BA}
\end{equation}

\subsection{Fine-tuning on Dense Passage Retrieval}
At the end of CoT-MAE pre-training, we only keep the encoder and discard the decoder.
The encoder weights are used to initialize a query encoder $f_q$ and a passage encoder $f_p$ for dense retrieval.
The query(or passage) encoder use the last layer's [CLS] embedding as the query(or passage) representation, denoted as $f_q(q)$(or $f_p(p)$).
The similarity of a query-passage pair $<q, p>$ is defined as an inner product:
$$s(q,p) = f_q(q) \cdot f_p(p)$$
Query and passage encoders are fine-tuned on the retrieval task's training corpus with a contrastive loss:
$$
\mathcal{L}=-\log \frac{\exp \left(s\left(q, p^{+}\right)\right)}{\exp \left(s\left(q, p^{+}\right)\right)+\sum_{l} \exp \left(s\left(q, p_{l}^{-}\right)\right)}
$$
, where $p^+$ denotes a positive passage and $\{p_l^{-}\}$ denotes a set of negative passages.
In practice, we reuse a widely adopted evaluation pipeline, i.e., Tevatron \cite{gao2022tevatron}.
The pipeline trains a first-stage-retriever using BM25 negatives, then trains a second-stage-retriever using BM25 negatives and hard negatives mined by the first-stage-retriever. The second-stage-retriever is used as the final retriever for evaluation.

\section{Experiments}

\begin{table*}[!htbp]
\centering
\small
\begin{tabular}{l|ccc|ccc|c}
\toprule  
& \multicolumn{3}{c|}{\textbf{MS-MARCO}} & \multicolumn{3}{c|}{\textbf{NQ}} & \textbf{TREC DL 20}\\
\textbf{Model} & MRR@10 & R@50 & R@1k & R@5 & R@20 & R@100 & nDCG@10 \\
\midrule
\midrule
\multicolumn{8}{c}{\textbf{Sparse retrieval}} \\
\midrule
BM25 & 18.7  & 59.2  & 85.7  & - & 59.1  & 73.7 & 47.9$\dagger$\\
docT5query \cite{nogueira2019doc2query}  & 21.5 & 64.4 & 89.1 & - & - & - & -  \\
DeepCT \cite{dai2019context} & 24.3  & 69.0  & 91.0  & - & - & - & -\\
GAR \cite{mao2020generation} & - & - & - & 60.9  & 74.4  & 85.3  & -\\
\midrule
\midrule
\multicolumn{8}{c}{\textbf{Dense retrieval}} \\
\midrule
ANCE \cite{xiong2020approximate} & 33.0  & - & 95.9  & - & 81.9  & 87.5 & - \\
SEED \cite{lu2021less} & 33.9  & - & 96.1  & - & 83.1  & 88.7  & -\\
TAS-B \cite{hofstatter2021efficiently} & 34.0  & - & 97.5  & - & - & - & \underline{69.3} \\
COIL \cite{gao2021coil} & 35.5  & - & 96.3  & - & - & - & -\\
ColBERT \cite{khattab2020colbert} & 36.0  & 82.9  & 96.8  & - & - & - & -\\
NPRINC \cite{lu2020neural} & 31.1  & - & 97.7  & 73.3  & 82.8  & 88.4  & - \\
Condenser \cite{gao2021condenser} & 36.6  & - & 97.4  & - & 83.2  & 88.4 & 66.5$\dagger$ \\
RocketQA \cite{qu2020rocketqa} & 37.0  & 85.5  & 97.9  & 74.0  & 82.7  & 88.5  & - \\
PAIR \cite{ren2021pair} & 37.9  & 86.4  & 98.2  & 74.9  & \underline{83.5}  & \underline{89.1}  & - \\
coCondenser \cite{gao2021unsupervised} & 38.2  & \underline{86.5}  & 98.4  & \textbf{75.8}  & \textbf{84.3}  & 89.0 & 68.0$\dagger$ \\
COSTA \cite{ma2022pre} & 36.6  & 84.1  & 97.3  & -  & -  & -  & 67.8$\dagger$ \\
RetroMAE \cite{liu2022retromae}$\star$ & \underline{39.3} & - & \underline{98.5}  & - & - & - & -\\
\bottomrule
CoT-MAE & \textbf{39.4} & \textbf{87.0} & \textbf{98.7} & \underline{75.5} & \textbf{84.3} & \textbf{89.3} & \textbf{70.4}\\
\bottomrule
\end{tabular}
\caption{Main results on the MS-MARCO passage ranking and Natural Questions (NQ) datasets. The best score on a given dataset is marked in \textbf{bold}, and the second best is \underline{underlined}. $\dagger$ denotes our reproduction with public checkpoints. $\star$ is a contemporaneous work.}
\label{table_cotmae}
\end{table*}

In this section, we first introduce the details of pre-training and fine-tuning. We then introduce the experimental results.
\subsection{Pre-training}
We initialize CoT-MAE's encoder from the pre-trained 12-layer BERT-\textit{base} and decoder from scratch.
Following \cite{gao2021unsupervised}, the pre-training dataset is constructed from the MS-MARCO passages corpus\footnote{ https://msmarco.blob.core.windows.net/msmarcoranking/msmarco-docs.tsv.gz} with 3.2M documents.
We use NLTK to split each document into sentences and group consecutive sentences into spans without exceeding the maximum span length equals 128. We use a uniform mixture of the three sampling strategies introduced, and dynamically select two spans in different epochs to form a span pair during the pre-training process.
We pre-train up to 1200k steps using AdamW optimizer, with a learning rate of 1e-4, and a linear schedule with warmup ratio 0.1. We train for 4 days with a global batch size of 1024 on 8 Tesla A100 GPUs. 
Due to the high compute budget in pre-training, we do not tune these hyperparameters, but leave that to future work.
After pre-training, we discard the decoder, only leaving the encoder for fine-tuning.

\subsection{Fine-tuning}
We fine-tune the pre-trained CoT-MAE on MS-MARCO passage ranking \cite{nguyen2016ms}, Natural Question \cite{kwiatkowski2019natural} and TREC Deep Learning (DL) Track 2020 \cite{craswell2020overview} tasks for evaluation.
Following coCondenser\cite{gao2021unsupervised}, we use the MS-MARCO corpus released in \cite{qu2020rocketqa}, 
following RocketQA\cite{qu2020rocketqa}, we use the NQ version created by DPR\cite{karpukhin2020dense}.
We reuse a widely adopt evaluation pipeline, i.e., Tevatron \cite{gao2022tevatron}, with a common fixed seed (42) to support reproducibility.
Note that, as we focus on improving the pre-training technique, we do NOT use any enhanced methods, such as distillation from a strong re-ranker or multi-vector representation, though they can lead to further improvements.
Following \cite{gao2021unsupervised, hofstatter2021efficiently}, for evaluation metrics, we use MRR@10, Recall@50, and Recall@1000 for MS-MARCO, Recall@5, Recall@20, Recall@100 for the NQ and nDCG@10 for TREC DL.

\paragraph{Baselines} 
Our baseline methods include the sparse retrieval method and the dense retrieval method, as shown in Table \ref{table_cotmae}.
Results of sparse retrieval baselines are mainly from \cite{qu2020rocketqa}, including BM25, docT5query \cite{nogueira2019doc2query}, DeepCT \cite{dai2019context} and GAR \cite{mao2020generation}.
Results of dense retrieval baselines are mainly from \cite{gao2021unsupervised, liu2022retromae, ren2021rocketqav2, ma2022pre}, including ANCE \cite{xiong2020approximate}, SEED \cite{lu2021less}, TAS-B \cite{hofstatter2021efficiently}, RetroMAE \cite{liu2022retromae}, and so on.

\subsection{Main Results}
We present the main results in Table \ref{table_cotmae}, which shows that CoT-MAE achieves considerable gains compared to competing baselines on three datasets.
On the MS-MARCO passage ranking dataset, CoT-MAE exceeds the previous state-of-the-art pre-training method coCondenser  1.2\% on MRR@10, which is a clear lead.
This suggests that the generative pre-training method is more effective than the contrastive learning method in leveraging context to enhance semantic representation modeling capacities.
Compared with methods like RocketQA that improve the fine-tuning stage or methods like ColBERT that adopt multi-vector, CoT-MAE also performs better.
On the NQ dataset, CoT-MAE also achieves comparable performance to coCondenser, outperforming the remaining methods. While coCondenser performs a little better on R@5, CoT-MAE outperforms it on R@100.
On the TREC DL dataset, CoT-MAE clearly outperforms the evaluated baselines and reaches 70 on nDCG@10.
In general, comparing CoT-MAE with the previous effective methods on the most commonly used benchmark datasets shows that the CoT-MAE pre-training process can effectively improve dense retrieval.
The improvement derives from two aspects. On the one hand, the pre-training method considers both the semantics of the tokens inside a text span and the semantical correlation between neighboring text spans.
On the other hand, the mixed data construction strategies and the asymmetric encoder-decoder structure with asymmetric masking strategies together contribute to efficient pre-training.

\section{Analysis}
We first compare CoT-MAE with the state-of-the-art distilled retrievers. Then we seek to understand the impact of different settings on CoT-MAE performance. Unless otherwise stated, our analyses are based on the MS-MARCO passage ranking task and pre-trained for 800k (not the 1200k in the main experiment) steps due to the high compute budget.

\subsection{Comparison with Distilled Retrievers}
As CoT-MAE focuses on improving pre-training to boost text representation in retrieval, comparison with retrievers distilled from re-rankers is optional and left here.
The typical passage ranking process involves a retrieval then re-ranking pipeline.
Re-ranker is a cross-encoder that models the full interaction between the query and the passage, which is strong but too computationally intensive to be applied in the retrieval.
\cite{ren2021rocketqav2, santhanam2021colbertv2, zhang2021adversarial} employ a re-ranker as a teacher to distill the retriever, trying to transfer the ability from the re-ranker to the retriever.

To further verify the effectiveness of CoT-MAE pre-training, we first compare the fine-tuned CoT-MAE retriever (without distillation) with the state-of-the-art distilled retrievers. Then we train a CoT-MAE reranker that reaches 43.3 on MRR@10 and use the re-ranker to distill the CoT-MAE retriever.

As shown in Table \ref{table_distill}, it is impressive that the fine-tuned CoT-MAE retriever (without distillation) achieves performance close to the distilled retrievers, demonstrating the effectiveness of CoT-MAE pre-training.
The fine-tuned CoT-MAE retriever slightly outperforms all distilled retrievers on R@1k, is only inferior to AR2 on R@50, and surpasses RocketQAv2 on MRR@10.
After applying distillation, the CoT-MAE retriever clearly exceeds all previously state-of-the-art distilled retrievers on MRR@10, R@50 and R@1k, showing that the CoT-MAE re-ranker can further improve the CoT-MAE retriever.

\begin{table}[!t]
\centering
\small
\begin{tabular}{l|c|ccc}
\toprule  
\textbf{Model}&  \textbf{Distilled} & MRR@10 & R@50 & R@1k\\
 \midrule
RocketQAv2 & \Checkmark & 38.8  & 86.2  & 98.1 \\
ColBERTv2 & \Checkmark & 39.7  & 86.8  & 98.4 \\
AR2 & \Checkmark & 39.5  & 87.8  & 98.6 \\
\midrule
CoT-MAE & \XSolidBrush & 39.2 & 87.2 & \textbf{98.7}\\
CoT-MAE & \Checkmark & \textbf{40.4} & \textbf{88.5} & \textbf{98.7}\\
\bottomrule
\end{tabular}
\caption{Comparison with distilled retrievers.}
\label{table_distill}
\end{table}

\begin{table}[!t]
\centering
\small
\begin{tabular}{cc|c||cc|c}
\toprule  
\textbf{Enc}& \textbf{Dec}& \textbf{MRR@10} & \textbf{Enc}& \textbf{Dec}& \textbf{MRR@10} \\
\midrule
0\% & 15\% & 37.8 & 45\% & 45\% & 39.0\\
0\% & 30\% & 38.4 & 45\% & 60\% & 38.8\\
\midrule
15\% & 15\% & 38.7 & 30\% & 15\% & 38.8 \\
15\% & 30\% & 38.8 & 30\% & 30\% & 38.9\\
15\% & 45\% & 38.9 & 30\% & 45\% & \textbf{39.2} \\
\bottomrule
\end{tabular}
\caption{Impact of mask rate. ``Enc'' denotes encoder, ``Dec'' denotes decoder.}
\label{table_mask_rate}
\end{table}

\subsection{Impact of Mask Rate}
\cite{wettig2022should} finds that a much larger mask ratio can outperform the 15\% baseline from BERT \cite{devlin2018bert}.
Therefore, we explore the effect of different mask rates for encoder and decoder.
As shown in Table \ref{table_mask_rate}, in our experiments, when the encoder mask rate equals 30\%, and the decoder mask rate equals 45\%, CoT-MAE achieves the best performance.
When the encoder mask rate does not exceed 30\%, the performance of CoT-MAE improves as the decoder mask rate increases.
When the encoder mask rate is as high as 45\%, the performance of CoT-MAE decreases slightly. We believe this is due to the insufficient context from the encoder when its mask rate is too large.
In general, CoT-MAE is quite robust to a wide range of mask rates, and an appropriate large mask rate can achieve relatively better performance, which is similar to \cite{wettig2022should}'s finding in BERT pre-training.

\subsection{Impact of Sampling Strategies}
In CoT-MAE‘s data construction process, we use three span sampling strategies, \textbf{Near}, \textbf{Olap} and \textbf{Rand}. We further explore the impact of different sampling strategies. We experiment with combinations of different strategies, covering using only a single strategy, mixing two strategies, and mixing three strategies. As shown in Table \ref{table_mask_strategy}, the best performance is achieved when mixing three strategies, with a slight drop when mixing two strategies and a larger drop when only a single strategy.
The trend shows that the diversity of data composition can benefit pre-training.
However, even with only a single strategy, CoT-MAE can also achieve a good result, further illustrating the effectiveness of the CoT-MAE model design.
\begin{table}[!t]
\centering
\small
\setlength{\tabcolsep}{4mm}{
\begin{tabular}{ccc|c}
\toprule  
\textbf{Near}& \textbf{Olap}& \textbf{Rand}& \textbf{MRR@10}  \\
\midrule
\Checkmark &  &  & 38.7 \\
 & \Checkmark &  & 38.7  \\
 &  & \Checkmark & 38.8  \\
\Checkmark & \Checkmark &  & 39.1  \\
\Checkmark &  & \Checkmark & 38.8 \\
 & \Checkmark & \Checkmark & 39.1  \\
\Checkmark & \Checkmark & \Checkmark & \textbf{39.2}   \\
\bottomrule
\end{tabular}}
\caption{Impact of sampling strategies. \Checkmark indicates that this sample strategy is selected.}
\label{table_mask_strategy}
\end{table}

\begin{table}[!t]
\centering
\small
\setlength{\tabcolsep}{2.5mm}{
\begin{tabular}{c|ccccc}
\toprule  
\textbf{Layers} & 1 & 2 & 3 & 4 & 6 \\
\midrule
\textbf{MRR@10} & 38.6 & \textbf{39.2} & 38.8 & 38.9 & 39.0 \\
\bottomrule
\end{tabular}}
\caption{Impact of decoder layers.}
\label{table_decoder_layer}
\end{table}



\begin{table*}[!htbp]
\centering
\small
\begin{tabular}{lcp{12cm}}
\toprule  
Relevant & Model  & Rank 1st passage\\
\midrule
\midrule
& & Query: what is operating system misconfiguration \\
\midrule
\multirow{2}{*}{\XSolidBrush} &\multirow{2}{*}{BERT$_{base}$} &  Passage: how to fix system drivers have stopped on my pc errors. windows operating system misconfiguration is the main cause of system drivers have stopped on my pc error codes... \\
 \midrule
\multirow{2}{*}{\XSolidBrush} &  \multirow{2}{*}{coCondenser}  & Passage: ne of the issues that users are experiencing on windows 10 is unexpected kernel mode trap error so letas see if we can fix this issue. unexpected kernel mode trap is a blue screen of death error that is caused... \\
 \midrule
\multirow{2}{*}{\Checkmark} & \multirow{2}{*}{CoT-MAE}  & Passage: windows operating system misconfiguration is the main cause of systemroot inf error codes. therefore, we strongly suggest using the downloadable systemroot inf repair kit to fix systemroot inf errors rakesh... \\
 \midrule
 \midrule
 & & Query: does cream of chicken soup have gluten \\
\midrule
\multirow{2}{*}{\XSolidBrush} &\multirow{2}{*}{BERT$_{base}$} &  Passage: directions. 1 whisk together the flour or cornstarch with the milk or milk substitute. 2 add the remaining ingredients and heat to a boil while whisking til fully dissolved and combined... \\
 \midrule
\multirow{2}{*}{\XSolidBrush} &  \multirow{2}{*}{coCondenser}  & Passage: stephanie1 0. i'm really new to being gluten - free, and i was wondering about campbell's soup. cream of mushroom and cream of chicken soup both contain modified food starch. i know that can be... \\
 \midrule
\multirow{2}{*}{\Checkmark} & \multirow{2}{*}{CoT-MAE}  & Passage: campbellas cream of chicken soup is not gluten - free. in fact, if you live in the us, there are no currently no campbellas soups that are gluten - free. \\
\bottomrule
\end{tabular}
\caption{Examples of rank 1st passage recalled by different models on the dev set of the MS-MARCO passage ranking dataset.}
\label{table_quality}
\end{table*}

\subsection{Impact of Decoder Layer Number}
We further explore the impact of different decoder layer numbers on CoT-MAE performance. We experiment on some most common layer settings, as shown in the table \ref{table_decoder_layer}. With a two-layer decoder, CoT-MAE achieves the best results in our experiment. A weaker or stronger decoder will decrease performance. When there is only one transformer layer in the decoder, the modeling ability of the decoder is too weak to fuse context embedding and unmasked token embeddings fully, resulting in inefficient utilization of information. 
When the number of layers is large, due to the stronger ability of the decoder, the masked auto-encoding task's dependence on the context embedding decreases, leading to insufficient constraints on encoder training.
In general, the performance of CoT-MAE for decoders with different layers is quite robust, and an appropriate deeper decoder can obtain relatively good performance.

\subsection{Impact of Decoder Weight Initialization}

\begin{figure}
\centering
\includegraphics[width=7.5cm]{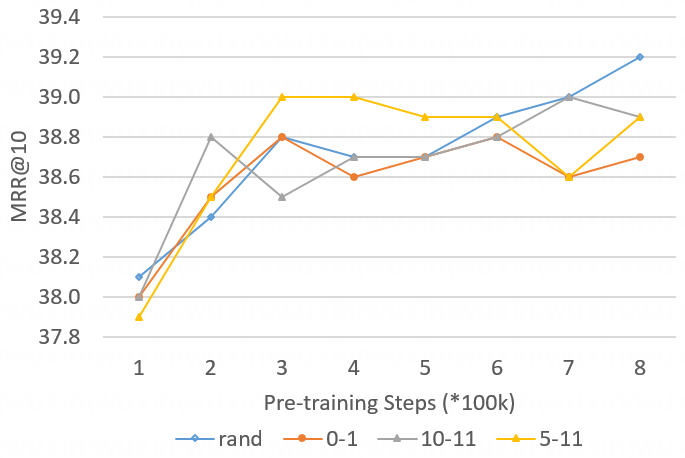}
\caption{Impact of decoder weight initialization. The $y$-axis shows the MRR@10 on MS-MARCO passage ranking.}
\label{pretrain_steps}
\end{figure}

The structure of CoT-MAE's encoder is the same as that of BERT, so the encoder is initialized directly with pre-trained BERT. 
In comparison, the decoder has only two layers, so different initialization options exist. We explore four different initialization methods: initialized from the first two layers (0,1), the uniform selected layers (5,11), the last two layers (10,11) of pre-trained BERT, and random initialization. We train 800k steps for each and plot their performance curves, as shown in Figure \ref{pretrain_steps}. 
In general, the long pre-training is sufficient for the decoder with limited parameters to converge, so the final results of the decoder are quite good.
After being pre-trained for more than 600k, random initialization continues to improve, while other initialization methods tend to fluctuate. It is worth noting that CoT-MAEs with different initializations significantly outperform coCondeser (38.2) when training only 200k, which further demonstrates the efficiency of CoT-MAE pre-training.

\subsection{Qualitative Analysis}
To qualitatively understand the gains from our proposed pre-training method, we show in Table \ref{table_quality} examples for which CoT-MAE can accurately recall relevant rank 1st passages. In the examples, although the rank 1st passage recalled by BERT or coCondenser has token-level overlap with the query, they are not highly relevant in semantics. In contrast, CoT-MAE performs better in semantic understanding due to the joint modeling of token and span context through masked auto-encoding in the pre-training stage. 
This further demonstrates that the CoT-MAE pre-training method is more effective than the previous pre-training methods.

\section{Conclusions and Future work}
This paper proposes a new generative pre-training method tailored for dense retrieval, considering the semantics of the tokens inside a text span and the semantical correlation between neighboring text spans. Experimental results show that CoT-MAE achieves considerable gains over competing baseline retrievers on benchmarks. 
In the future, we will further explore the possibility of applying contextual masked auto-encoder in multilingual and multimodal domains.

\bibliography{aaai23}

\end{document}